\documentclass[11pt,a4paper]{article}
\usepackage[hyperref]{emnlp-ijcnlp-2019}
\usepackage{times}
\usepackage{latexsym}

\usepackage{amsmath,amsfonts,amssymb,amsthm,comment,natbib}
\usepackage{mathtools}
\usepackage{todonotes}
\usepackage{url}
\usepackage{float}
\usepackage{algorithm2e}
\usepackage{commath,booktabs}

\aclfinalcopy 

\title{ Semantic Role Labeling with Iterative Structure Refinement}

\author{ {Chunchuan Lyu}$^1$  ~~ {Shay B. Cohen}$^1$ ~~ {Ivan Titov}$^{1,2}$ 
 \smallskip \\
      {ILCC, School of Informatics, University of Edinburgh} $^1$  \\
{ILLC, University of Amsterdam}$^2$ 
 \smallskip \\
 {\tt chunchuan.lv@gmail.com} ~~ {\tt scohen@inf.ed.ac.uk} ~~ {\tt ititov@inf.ed.ac.uk}  \\
}

\begin{document}
\maketitle 
\begin{abstract}
 
 Modern state-of-the-art Semantic Role Labeling (SRL) methods rely on expressive sentence encoders (e.g., multi-layer LSTMs) but tend to model only local (if any) interactions between individual argument labeling decisions. This contrasts with earlier work and also with the intuition that the labels of individual arguments are strongly interdependent.
We model interactions between argument labeling decisions through {\it iterative refinement}. Starting with an output produced by a factorized model, we iteratively refine it using a refinement network. Instead of modeling arbitrary interactions among roles and words, we encode
prior knowledge about the SRL problem by designing a restricted network architecture capturing non-local interactions. This modeling choice prevents overfitting and results in an effective model, outperforming strong factorized baseline models on all 7 CoNLL-2009 languages, and achieving state-of-the-art results on 5 of them, including English. ~\footnote{To appear in EMNLP 2019} 

\end{abstract}
\section{Introduction}
\begin{figure}[t!]
\centering
\includegraphics[width=1\columnwidth]{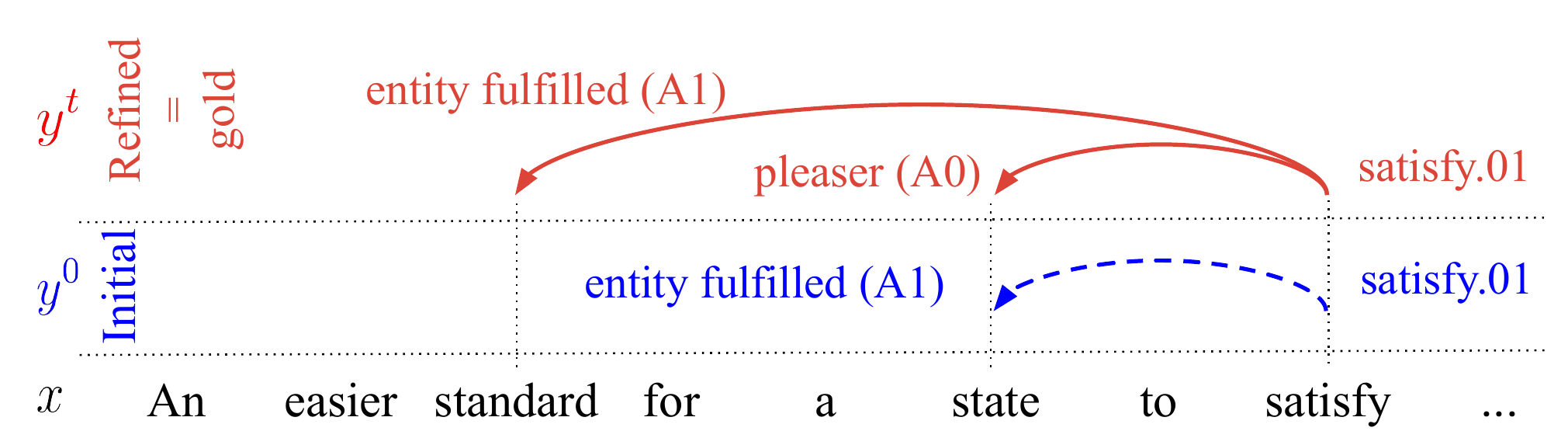}
\caption{An example of structured refinement, the sentence fragment is from CoNLL-2009: the initial prediction by the factorized model in blue, the refined one (identical to the gold standard) in red.}
\label{fig:refine-example}
\end{figure}

Semantic role labeling (SRL), originally introduced by \citet{Gildea2000AutomaticLO}, 
involves the prediction of predicate-argument structure, i.e., identification of
arguments and their assignment to underlying \textit{semantic roles}. 
Semantic-role representations have been shown to be beneficial in many NLP applications, including question answering~\cite{DBLP:conf/emnlp/ShenL07}, information extraction \cite{DBLP:conf/kcap/ChristensenMSE11} and machine translation~\cite{Marcheggiani2018ExploitingSI}. In this work, we focus on dependency-based SRL~\cite{conll2009}, a popular version of the task which involves identifying syntactic heads of arguments rather than marking entire argument spans (see the graph in red in Figure~\ref{fig:refine-example}). Edges in the dependency graphs are annotated with semantic roles (e.g., {\sc A0:pleaser}) and the predicates are labeled with their senses from a given sense inventory (e.g., {\sc satisfy.01} in the example).

Before the rise of deep learning methods, the most accurate SRL methods relied on modeling high-order interactions in the output space (e.g., between arguments or arguments and predicates)~\cite{Watanabe2010ASM,Toutanova2008AGJ}. Earlier neural methods can model such output interactions through a transition system, and achieve competitive performance~\cite{henderson-etal-2013-multilingual}. However, current state-of-the-art SRL systems use powerful sentence encoders (e.g., layers of LSTMs~\cite{Li2018DependencyOS,P17-1044} or  multi-head self-attention~\cite{Strubell2018LinguisticallyInformedSF}) and factorize over small fragments of the predicted structures. Specifically, most modern models process individual arguments and perform predicate disambiguation independently. 
The trend towards more factorizable models is not unique to dependency-based SRL but common for most structured prediction tasks in NLP~\cite{Kiperwasser2016SimpleAA,DBLP:conf/iclr/DozatM17,dozat-manning-2018-simpler}. The only major exception is language generation tasks, especially machine translation and language modeling, where
larger amounts of text are typically used in training.


Powerful encoders, in principle, can capture long-distance dependencies and hence alleviate the need for modeling high-order interactions in the output. However, capturing these interactions in the encoder would require substantial amounts of data. Even if we have domain knowledge
about likely interactions between components of the predicted graphs, it is hard to inject this knowledge in an encoder. 

Consider the example in Figure~\ref{fig:refine-example}. The argument {\it `state'} appears in the highly ambiguous syntactic position '[.] {\it to satisfy}'. All three core semantic roles of the predicate {\sc satisfy.01} can in principle appear here: patient ({\sc A1:entity$\_$fulfilled}, as in `{\it a sweet tooth to satisfy'}), instrument ({\sc A2:method}, as in `{\it a little dessert to satisfy your sweet tooth}') and agent (\textsc{A0: pleaser}, as in our actual example). The basic factorized model got it wrong, assigning \textsc{A1} to the argument {\it `state'}. However, taking into account other arguments, the model can correct the label. The configuration `{\sc A1} {\it to satisfy}' is more likely when an agent ({\sc A0}) is present in the sentence. The lack of an agent  boosts the score for the correct configuration `{\sc A0} {\it to satisfy}'.

Our {\it iterative refinement} approach encodes the above intuition.
In iterative refinement~\cite{Lee2018DeterministicNN}, a refinement network repeatedly takes previous output as input and produces its refined version. Formally, we have $$y^{t+1} =  \mathrm{Refine}(x,y^t).$$ Naturally, such refinement strategy also requires an initial prediction $y^0$,  which is produced by a (`base') factorized model. 

Refinement strategies have been successful in machine translation~\cite{Lee2018DeterministicNN,Novak2017IterativeRF,Xia2017DeliberationNS,Hassan2018AchievingHP}, but their effectiveness in other NLP tasks 
is yet to be demonstrated.\footnote{See extra discussion and related work in section~\ref{sec:related}.} We conjecture that this discrepancy is due to differences in data availability. Given larger amounts of training data typically used in machine translation, their base models and refinement networks overfit to a lesser extent. Overfitting  in (1) the base model and (2) the refinement network are both problematic. The first implies that either there are no mistakes in the base models in the training set or their distribution is very different from that in the test regime, so the training material for the inference networks ends up being misleading. The second naturally means that refinement will fail at test time. We address both these issues by designing restricted inference networks and adding a specific form of noise when training them.


Our structured refinement network is simple but encodes non-local dependencies.
Specifically, it takes into account the information about the role distributions on the previous iteration aggregated over the entire sentences but not the information what the other arguments are.
It is a coarse compressed representation of the prediction, yet it represents long-distance information not readily available within the factorized base model. While this is not the only possible design, we believe that the empirical gains from using this simple refinement network, demonstrate the viability of our general framework of iterative refinement with restricted inference networks.
They also suggest that intuitions underlying declarative constraints used in early work on SRL~\cite{Punyakanok:2008:ISP:1403157.1403162,das2012exact} can be revisited but now encoded in a flexible soft way to provide induction biases for the refinement networks. We leave this for future work.



We consider the CoNLL-2009 dataset~\cite{conll2009}. We start with a strong factorized baseline model, which already achieves state-of-the-art results on a subset of the languages. 
Then, using our structure refinement network, we improve on this baseline on all 7 CoNLL-2009 languages. The model achieves best-reported results in 5 languages, including English.  We also observe improvements on out-of-domain test sets, confirming the robustness of our approach. We perform experiments demonstrating the importance of adding noise, and ablation studies showing the necessity of incorporating output interactions. Furthermore, we provide analysis on constraint violations and errors on the English test set.\footnote{ The code and experiment settings {\bf will} be accessed at \url{ https://github.com/ChunchuanLv/Iterative_Inference}}


\section{Related Work}\label{sec:related}


Learning to refine predictions from neural structured prediction models has recently received significant attention.
Our approach bears similarity to methods used in machine translation~\cite{Lee2018DeterministicNN,Novak2017IterativeRF,Xia2017DeliberationNS}. All these methods refine a translated sentence produced by a seq2seq model with another seq2seq model. Among them, the deliberation networks  by ~\newcite{Xia2017DeliberationNS} rely on BiLSTMs and improve initial predictions from an competitive baseline and obtain state-of-art-results on English-to-French translation. Later, it has been shown that the deliberation networks can improve translation when used within the Transformer framework~\cite{Hassan2018AchievingHP}.

Certain approaches, not necessarily directly optimized for refinement, can nevertheless be regarded as iterative refinement methods.
 Structured prediction energy networks (SPENs) are trained to assign global energy scores to output structures, and the gradient descent is used during inference to minimize the global energy~\cite{SPEN}. As the gradient descent involves iterative optimization, its steps can be viewed as iterative refinement. In particular, \newcite{Belanger2017EndtoEndLF} build a SPEN for SRL, but for the span-based formalism, not the dependency one we consider in this work. While they improve over their baseline model, their baseline model used multilayer perceptron to encode local factors, thus the encoder power is limited. Moreover their refined model performs worse in the out-of-domain setting than their baseline model, indicating overfitting~\cite{Belanger2017EndtoEndLF}.

In the follow-up work, \newcite{Tu2018LearningAI,Tu2019BenchmarkingAI} introduce inference networks to replace gradient descent. Their inference networks directly refine the output. 
Improvements over competitive baselines are reported on part-of-speech tagging, named entity recognition and CCG super-tagging~\cite{Tu2019BenchmarkingAI}. However, their inference networks are distilling knowledge from a tractable linear-chain conditional random field (CRF) model. Thus, these methods do not provide direct performance gains. More importantly, the interactions captured in these models are likely local, as they learn to mimic Markov CRFs.

Denoising autoencoders ~\cite{Vincent2008ExtractingAC} can also be used to refine structure. Indeed, image segmentation can be improved through iterative inference with denoising autoencoders~\cite{CDAE_Image,DROZDZAL20181}. Their framework is very similar to ours, albeit we are working in a discrete domain. One other difference is that by using a convolutional architecture in the refinement network, they are still modeling only local interactions.  
At a more conceptual level, ~\newcite{GDAE} argued that a denoising autoencoder should not be too robust to the input variations as to ignore the input. 
This indicates that we should not expect refinement networks to correct all the errors, even in theory, and hence, the refinement networks do not need to be particularly powerful.  


Very recently, ~\newcite{Wang2019SecondOrderSD} used high order statistical model for Semantic Dependency Parsing~\cite{oepen-etal-2015-semeval}, and obtain improvements over strong baseline using BiLSTM. They attempted loopy belief propagation and mean field variational inference for inference, and train the model end to end. Such inference steps are well motivated. This work is similar to energy network approach~\cite{SPEN}, while a global score function is provided, and approximate inference steps are used. Comparing to ours, the inference can also be regarded as iterative structure refinement. Yet, we do not provide a global score and directly try to model the refinement. In principle, our formalization should give us more liberty in terms of designing the refinement network. 

\section{Dependency Semantic Role Labeling}
In this section, we introduce the notation and present
 our factorized baseline model. 

\subsection{Notation}
In dependency SRL, for each sentence of length $n$, we have a sequence of words $ \mathrm{w}$, dependency labels $\mathrm{dep}$, part-of-speech tags $ \mathrm{pos}$, each being a discrete sequence of length  $n$.  To simplify notation, we consider one predicate at a time. We denote the number of roles by $r$, it  includes the `null' role, signifying that the corresponding word is not an argument of the predicate. Formally, $\mathrm{P}  \in \Delta_{m-1}$ is the probability distribution over $m$ predicate senses, and $\Delta_{m-1}$ represents the corresponding probability simplex. We also have predicate sense embeddings $\Pi  \in \mathbb{R}^{m \times d_\pi }$, and index $j$, throughout the discussion, refers to the position of the target predicate in the sentence. $\mathrm{R} \in \Delta^n_{r-1}$ is a matrix of size $n \times r$ such that each row sums to 1, corresponding to a probability distribution over roles. In particular  $\mathrm{R}_{i,0}$ is the probability of $i$-th word not being an argument of the predicate. 

We index role label and sense predictions from different refinement iterations (`time steps') with $t$, i.e. $\mathrm{P}^t$ and  $\mathrm{R}^t$. The index $t$ ranges from $0$ to $T$, and $\mathrm{P}^0$ and  $\mathrm{R}^0$ denotes the predictions from the factorized baseline model.  Details (e.g., hyperparameters) are provided in the appendix. 

\subsection{Factorized Model } \label{sec:base}
Similarly to recent approaches to SRL and semantic graph parsing~\cite{P17-1044,Li2018DependencyOS,dozat-manning-2018-simpler}, our factorized baseline model starts with concatenated embeddings~$x$. Then, we encode the sentence with a BiLSTM, further extract features with an MLP (multilayer perceptron) and apply a bi-affine classifier to the resulting features to label the words with roles.  
We also use a predicate-dependent linear layer for sense disambiguation. 

More formally, we start with getting a sentence representation by concatenating embeddings. We have  $x^{ \mathrm{w}} \in \mathbb{R}^{n \times d_w}$, $x^{ \mathrm{dep} } \in \mathbb{R}^{n \times d_{\delta}}$, $x^{ \mathrm{pos}} \in \mathbb{R}^{n \times d_p}$ for words, dependency labels and part-of-speech tags, respectively. 
We concatenate them to form a sentence representation:
\begin{align}
x &= x^{ \mathrm{w}}\circ x^{ \mathrm{dep}} \circ x^{ \mathrm{pos}}  \in \mathbb{R}^{n \times d_x}
\end{align}
\noindent We further encode the sentence with a BiLSTM:
\begin{align}
 \mathrm{h} = & \text{BiLSTM}(x) \in  \mathbb{R}^{n\times d_{h}} 
\end{align}
\noindent From these context-aware word representations, we produce features for argument identification and role labeling that will be used by a bi-affine classifier. Note that, for every potential predicate-argument dependency (i.e. a candidate edge), we need to produce representations of both endpoints: the argument and the predicate `sides'.
For the argument side,  $\mathrm{h}^{\rho_0}$ will be used to compute the logits for argument identification and $ \mathrm{h}^{\rho_1}$ will be used for deciding on its role:
\begin{align}
 \mathrm{h}^{\rho_0} = & \text{MLP}(\mathrm{h} )  \in \mathbb{R}^{n\times d_{\rho_0}} \\
 \mathrm{h}^{\rho_1} = & \text{MLP}(\mathrm{h} )  \in \mathbb{R}^{n\times d_{\rho_1}} 
\end{align}
\noindent Similarly, for the predicate side, we also extract two representations $ \mathrm{h}^\varrho_0$ and $\mathrm{h}^{\varrho_1} $ (recall that the predicate is at position~$j$):
\begin{align}
 \mathrm{h}^{\varrho_0} = & \text{MLP}(\mathrm{h}_j) \in \mathbb{R}^{ d_{\rho_0}} \\
 \mathrm{h}^{\varrho_1} = & \text{MLP}(\mathrm{h}_j) \in \mathbb{R}^{ d_{\rho_1}} 
\end{align}
\noindent We then obtain logits $\mathrm{I}^{\rho_0}$ corresponding to the decision to label arguments as {\it null}, and logits $\mathrm{I}^{\rho_1}$ for other roles.
 So, we have:
\begin{align}
\mathrm{I}^{\rho_0} = &\text{BiAffine}( \mathrm{h}^{\varrho_0},\mathrm{h}^{\rho_0}) \in \mathbb{R}^{n} \label{eq:init_null}\\ 
\mathrm{I}^{\rho_1} = &\text{BiAffine}( \mathrm{h}^{{\varrho_1}},\mathrm{h}^{\rho_1}) \in \mathbb{R}^{n\times (r-1)} \label{eq:init_other}
\end{align}
\noindent Unlike \newcite{dozat-manning-2018-simpler}, where argument identification and role labeling are trained with two losses,\footnote{The separate processing of $\mathrm{I}^{\rho_0} $ and $\mathrm{I}^{\rho_1} $ rather than using a single MLP for all roles, including {\it null}, results in extra representation power allocated for the argument identification subtask.} 
 we feed them together into a single softmax layer to compute the semantic-role distribution $\mathrm{R}^0$:
\begin{align}
\mathrm{I}^{\alpha} = & \mathrm{I}^{\rho_0} \circ \mathrm{I}^{\rho_1} \in \mathbb{R}^{n\times r} \label{eq:initR}\\ 
\mathrm{R}^0 =& \mathrm{Softmax}(\mathrm{I}^{\alpha})  \in \Delta^n_{ r-1}
\end{align}
\noindent Now, for sense disambiguation, we need to extract yet another predicate representation $\mathrm{h}^\pi$:
\begin{align}
 \mathrm{h}^\pi = & \text{MLP}(\mathrm{h}_j) \in \mathbb{R}^{d_\pi}
\end{align}
\noindent In the formalism we use  (PropBank), senses are predicate-specific, so we use predicate-specific sense embedding matrices $\Pi$. The matrix $\Pi$ acts as a linear layer before softmax:
\begin{align}
\mathrm{I}^{\pi} = & \Pi \cdot  \mathrm{h}^{\pi }  \in \mathbb{R}^{m} \label{eq:initP}\\
\mathrm{P}^0 =& \mathrm{Softmax}(\mathrm{I}^{\pi})  \in \Delta_{m-1}
\end{align}
This ends the description of  our baseline model, which we also use to get initial predictions for iterative refinement.

\section{Structured Refinement Network  } \label{sec:SRN}
In this section, we introduce the structured refinement network for dependency SRL. When doing refinement, it has access to the  roles distribution $\mathrm{R}^t \in  \Delta^n_{r-1}$ and the sense distribution $\mathrm{P}^t \in  \Delta_{ m -1}$ computed at the previous iteration (i.e. time $t$). In addition, it exploits the sentence representation $x\in \mathbb{R}^{ n\times d_x}$. Our refinement network is limited and structured, in the sense that it only has access to a compressed version of the previous prediction, and the network itself is a simple MLP. 

Similarly to our baseline model, we extract feature vectors $ \mathrm{g}$ from input $x$ and further separately encode the argument representation $ \mathrm{g}^\alpha $ and the predicate token representation  $\mathrm{g}^\pi $:
\begin{align}
 \mathrm{g} = & \text{BiLSTM}(x) \in  \mathbb{R}^{n\times d_{h}}\\ 
 \mathrm{g}^\alpha = & \text{MLP}(\mathrm{g} ) \in \mathbb{R}^{n\times d_g}\\
 \mathrm{g}^\pi = & \text{MLP}(\mathrm{g}_j) \in \mathbb{R}^{ d_{g}}
\end{align}
\noindent To simplify the notation, we omit indexing them by $t$, except for $\mathrm{R}^t$ and $\mathrm{P}^t$. We use two refinement networks, one for roles and another one for predicate senses.

\subsection{Role Refinement Network}
First, we describe our structured refinement network for role labeling. We use $i$ to index arguments. We obtain a compressed representation $\mathrm{o}_{i}$ used for refining $\mathrm{R}^t_i$ by summing up the  probability mass for all roles, excluding the null role:
\begin{align}
\mathrm{o}_{i,u} &=  \underset{k\neq i}{\sum}\mathrm{R}^t_{k,u} \in \mathbb{R}\\
\mathrm{o}_{i}  & = [\mathrm{o}_{i,u}] \in \mathbb{R}^{ r-1}
\end{align}
\noindent Intuitively, $\mathrm{o}_{i}$ is the aggregation of all other roles being labeled by the current predicate. We concatenate $\mathrm{o}_{i}$ with feature vectors of the current argument $ \mathrm{g}^\alpha $, predicate  $\mathrm{g}^\pi $, the relaxed predicate sense embedding $\Pi^\intercal \cdot \mathrm{P}^t $ and the role probability itself ($\mathrm{R}_{i}$) to form the input to a two-layer network:
\begin{align}
 \mathrm{z}^{\alpha}_i = &\mathrm{R}^t_{i}\circ \mathrm{o}_{i} \circ   \mathrm{g}^{\alpha}_i  \circ  \mathrm{g} ^\pi  \circ (\Pi^\intercal \cdot \mathrm{P}^t) \label{eq:role_rep}\\
   \mathrm{z}^{\alpha}_i  \in& \mathbb{R}^{ 2r-1+2d_g+d_\pi}  \nonumber\\ 
   \label{eq:m_rep}
\mathrm{M}^{\alpha}_{i} = &  W^\alpha  \cdot  \sigma (W_\alpha  \cdot  \mathrm{z}^{\alpha}_i )    \in \mathbb{R}^r \\
\mathrm{M}^{\alpha} =& [\mathrm{M}^{\alpha}_{i,u}] \in  \mathbb{R} ^{n \times r},
\end{align}
\noindent where $\sigma$ is the logistic sigmoid function,  $W_\alpha \in \mathbb{R}^{ d_r \times (2r-1+2d_g+d_\pi ) },W^\alpha \in \mathbb{R}^{ r \times d_r}$ are learned linear mappings. We obtain our refined logits $\mathrm{M}^{\alpha}_i$ for the $i$-th argument; $\mathrm{M}^{\alpha}$ refers to the stacked matrix of logits for all arguments. To obtain the refined role distribution, we add up $\mathrm{M}^{\alpha}$ and $\mathrm{I}^{\alpha}$ that we got from the baseline model, and follow that by a softmax layer:
\begin{align}
   \mathrm{R}^{t+1} = &  \mathrm{Softmax}(\mathrm{M}^{\alpha}+\mathrm{I}^{\alpha}) \in  \Delta^n_{  r-1}
\end{align}

\subsection{Sense Refinement Network}
To build a representation for sense disambiguation, we simply compute the probability mass for each role (excluding the null role)  to obtain $\mathrm{r}^{\pi}$, and concatenate this with  $\mathrm{g}^\pi $ and $\Pi^\intercal \cdot \mathrm{P}^t $:
\begin{align}
\mathrm{r}^{\pi} = & \underset{k}{\sum}\mathrm{R}^t_{k,1:}  \in \mathbb{R}^{ r-1} \\
  \mathrm{z}^{\pi} = &( \Pi^\intercal \cdot \mathrm{P}^t) \circ \mathrm{r}^\pi  \circ   \mathrm{g}^\pi \in \mathbb{R}^{ r-1+d_g+d_\pi}  \label{eq:pre_rep}
\end{align}
\noindent Differently from the role refinement network, sense prediction is predicate-specific. Therefore, we first map $ \mathrm{z}^{\pi}$ to $\mathbb{R}^{d_\pi}$, and then take the inner product with the predicate-specific sense embeddings $\Pi$ to get the refined logits:
\begin{align}
\mathrm{M}^{\pi} = &  \Pi  \cdot  W^\pi   \cdot  \sigma (W_\pi  \cdot  \mathrm{z}^{\pi} )    \in \mathbb{R} ^m  \label{eq:m_p_rep}
\end{align}
\noindent Similarly to role refinement,  $\sigma$ is the logistic function, 
$W_\pi \in \mathbb{R}^{ d_r \times (r-1+d_g+d_\pi ) },W^\pi \in \mathbb{R}^{ m \times d_r}$ are learned linear mappings. Again, we combine the logits $\mathrm{M}^{\pi}$ and $\mathrm{I}^{\pi}$ before the softmax layer:
\begin{align}
   \mathrm{P}^{t+1} = &  \mathrm{Softmax}(\mathrm{M}^{\pi}+\mathrm{I}^{\pi}) \in  \Delta_{ m-1} 
\end{align}


\subsection{Weight Tying}

Our refinement networks are similar to the denoising autoencoders (DAEs; \citealt{Vincent2008ExtractingAC}), so we use the weight-tying technique popular with DAEs. 
We believe that the technique may be even more effective here as the amount of labeled data for SRL is lower than in many usual applications of DAEs.
We  tie $W_\alpha$ with a subset of $W^\alpha$ rows: specifically with the rows acting on $R^t_i$ in the computation of $M_i^\alpha$ (see equations~\ref{eq:role_rep} and~\ref{eq:m_rep}). Similarly, we tie  $W_\pi$ with the part of $W^\pi$  corresponding to $ \Pi^\intercal \cdot \mathrm{P}^t$ (see equations \ref{eq:pre_rep} and \ref{eq:m_p_rep}). Formally,
\begin{align}
 W^\alpha &= W_\alpha^\intercal [:r] \\
 W^\pi &= W_\pi^\intercal [:d_\pi]
\end{align}
\noindent where $W[:k]$ takes the first $k$ rows of matrix $W$.  

\subsection{Self Refinement}
We describe a simpler version of the refinement network which we will use in experiments to test whether the improvements with the structured refinement network
over the factorized baseline are genuinely coming 
from modeling interaction between arguments rather than from simply combining multiple classifiers. This simpler refinement network
does not account for any interactions between arguments. Instead of equations~\ref{eq:role_rep} and ~\ref{eq:pre_rep}, we have:
\begin{align}
 \mathrm{z}^{\alpha}_i = &\mathrm{R}^t_{i}\circ  \mathrm{g}^{\alpha}_i  \circ  \mathrm{g} ^\pi \in \mathbb{R}^{ r+2d_g+d_\pi}  \label{eq:role_rep-}  \\
  \mathrm{z}^{\pi} = &( \Pi^\intercal \cdot \mathrm{P}^t)   \circ   \mathrm{g}^\pi \in \mathbb{R}^{ d_g+d_\pi}  \label{eq:pre_rep-} 
\end{align}
Everything else is kept the same as in the full model, expect that the size of $W^\alpha$ and $W^\pi$ needs to be adjusted. We refer to this ablated network as the {\it self-refinement network}. 

\section{Training for Iterative Structure Refinement}
In this section, we describe our training procedure.

\subsection{Two-Stage Training}
We have two models: the baseline model, producing the initial predictions, and the iterative refinement network, correcting them.
While it is possible to train them jointly, we find joint training slow to converge.
Instead, we train the factorized baseline model first and then optimize the refinement networks while keeping the baseline model fixed.

\subsection{Stochastic Training}

Our baseline model overfits to the training set, and, if simply training on its output, our refinement network would learn to copy the base predictions.  Instead, we perturb the baseline prediction during training. Naturally, we can add dropout~\cite{JMLR:v15:srivastava14a} and recurrent dropout~\cite{Gal2016ATG} to our neural networks. However, for the smaller data set we use, we find this not sufficient. 
In particular, we use Gumbel-Softmax instead of Softmax. $\mathrm{Gumbel}\mbox{-}\mathrm{Softmax}(\mathrm{I}) = \mathrm{Softmax}(\mathrm{I}+\lambda_{g}\epsilon)$, where the random variable $\epsilon$ is drawn from the  standard Gumbel distribution~\cite{Maddison2016TheCD,Jang2016CategoricalRW},  
and $\lambda_{g}$ is a hyper-parameter controlling decoding stochasticity.\footnote{A more canonical way of controlling stochasticity is to use the temperature but we prefer not to scale the gradient.} 

\begin{table*}[h!]
\centering
 \begin{tabular}{c| c c c c c c c | c} 
 \hline
 Model &Ca &Cz& De& En& Ja & Es & Zh & Avg.\\ [0.5ex] 
 \hline
 \newcite{roth-lapata-2016-neural}  & -&- & \bf80.10 &86.7 & -&80.20 &79.4 & -\\ 
 \newcite{Marcheggiani2017ASA}  & - & 86.00& -& 87.7 &-&  80.30 & 81.2 & -\\ 
 \newcite{mulcaire-etal-2018-polyglot}*  &79.45 & 85.14&69.97& 87.24 &76.00 & 77.32 &  81.89 & 79.57\\ 
 Previous best single model & 80.32 & 86.02& \bf80.10& 90.40 &78.69& 80.50 & \bf 84.30 & 82.90 \\ 
 \hline
 Baseline model &  80.69   & 87.30  &  75.06   &      90.65  &    81.97 & 79.87 & 83.26 & 82.69 \\
 Structured refinement  & {\bf{  80.91 }}  & \bf{ 87.62 } & 75.87 & \bf 90.99   &   {\bf  82.54 } &   {\bf 80.53} &  83.31 & \bf 83.11 \\
 \hline
 \end{tabular}
\caption {Labeled F1 score (including senses) for all languages on the CoNLL-2009 in-domain test set. For previous best result, Catalan is from~\newcite{Zhao2009MultilingualDL}, Japanese is from~\newcite{Watanabe2010ASM}, Czech is from~\newcite{henderson-etal-2013-multilingual}, German and Spanish are from~\newcite{roth-lapata-2016-neural}, English is from~\newcite{Li2018DependencyOS} and Chinese is from~\newcite{Cai2018AFE}. We report the best testing results from~\newcite{mulcaire-etal-2018-polyglot}. }
 \label{tb:mul}
\end{table*}

\subsection{Loss for Iterative Refinement}

Let us denote gold-standard labels for roles and predicates as
 $\mathrm{R}^*$ and $\mathrm{P}^*$. We use two separate losses $\mathcal{L}_{\mathrm{base}}(\mathrm{R}^*,\mathrm{P}^*,x)$ and $\mathcal{L}_{\mathrm{refine}}(\mathrm{R}^*,\mathrm{P}^*,x) $ for our two-stage training. We define losses for predictions from each refinement iteration and sum them up:
\begin{align}
\mathcal{L}_{\mathrm{base}}(\mathrm{R}^*,\mathrm{P}^*,x) &= \mathcal{L}(\mathrm{R}^*,\mathrm{R}^0)+\mathcal{L}(\mathrm{P}^*,\mathrm{P}^0)\\
\mathcal{L}_{\mathrm{refine}}(\mathrm{R}^*,\mathrm{P}^*,x) &= \sum_{t=1}^T \mathcal{L}(\mathrm{R}^*,\mathrm{R}^t)+\mathcal{L}(\mathrm{P}^*,\mathrm{P}^t) 
\end{align}
\noindent 
We adopt the Softmax-Margin loss~\cite{gimpel-smith-2010-softmax,blondel2019learning} for individual $\mathcal{L}$. Effectively, we subtract 1 from the logit of the gold label, and apply the cross entropy loss. 
\section{Experiments}
\noindent {\bf Datasets }We conduct experiments on CoNLL-2009~\cite{conll2009} data set for all languages, including Catalan (Ca), Chinese (Zh), Czech (Cz), English (En), German (De), Japanese (Jp) and Spanish (Es).  We use the  predicted part-of-speech tags, dependency labels, and pre-identified predicate, provided with the dataset.  
The statistics of datasets are shown in Table~\ref{tb:stats}.
\begin{table}[ht!]
\centering
\begin{tabular}{c | c c c}
 & \#sent & \#pred & \#pred/\#sent \\ 
 \hline
 Ca  & 13200 & 37444 & 2.84 \\
 Cz & 38727 & 414133 & 10.69 \\
 De & 36020 & 17400 &0.48 \\
 En &39279 & 179014 &4.56\\
 Ja & 4393 &25712 &5.85\\
 Es & 14329 &43828&3.06\\
 Zh & 22277 & 102827&4.62\\
 \hline
\end{tabular}
\caption{Number of sentences and predicates in training set of different languages.}
\label{tb:stats}
\end{table}

\begin{table*}[ht!]
\centering
 \begin{tabular}{c| c c  c c c c c | c} 
 \hline
 Model  &Ca & Cz &De  & En&  Ja& Es & Zh & Avg.\\ [0.5ex] 
 \hline
 Baseline  & 81.69  & 88.43 & 73.97   &  89.60 &      82.96 & 80.49 & 85.27 &83.20 \\
 \hline
Full   &\bf 82.11  &88.62 & 74.95 & 89.82    & \bf  83.60   & \bf  81.19 & \bf  85.52 &\bf 83.69 \\
1 iteration  & 82.07 & 88.62  &  \bf  75.07& 89.93  & 83.49  & 81.03  & 85.47 & 83.40  \\
 un-tied   & 81.99   &88.61   & 75.04  &  89.79   & 83.47  & 80.89      & 85.49 &83.61  \\
 no Gumbel   &  82.07    &\bf 88.71  &  74.62   & \bf 90.08   & 83.33   & 80.55     & 85.42   &   83.54\\
 \hline
 \end{tabular}
\caption {Labeled F1 score (including senses)  for all languages on development set for different configurations.}
 \label{tb:ab1}
\end{table*}
\noindent {\bf Hyperparameters} 
We use ELMo~\cite{Peters2018DeepCW} for English, and FastText embeddings~\cite{bojanowski2017enriching,grave-etal-2018-learning} for all other languages. 
We train and run the refinement networks for two iterations. All other hyper-parameters are the same for all languages, except BiLSTMs for English is larger than others.  

\noindent {\bf Training Details}
Training the refinement network takes roughly 2 times more time than the baseline models, as it requires running BiLSTMs. 
The extra computation for the structured refinement network is minimal. For English, training the iterative refinement model for 1 epoch takes about 6 minutes on one 1080ti GPU. Adam is used as the optimizer~\cite{Kingma2015AdamAM}, with the learning rate of 3e-4. We use early stopping on the development set.  We run 600 epochs for all baseline models, and 300 epochs for the refinement networks. Batch sizes are chosen from 32, 64, or 128 to maximize GPU memory usage. Our implementation is based on PyTorch and AllenNLP~\cite{paszke2017automatic,Gardner2018AllenNLPAD}. 
\subsection{Results and Discussions}
\noindent{\bf Test Results} Results for all CoNLL-2009 languages  on the standard (in-domain) datasets are presented in Table~\ref{tb:mul}. We compare our best model to the best previous single model for the corresponding language (excluding ensemble ones). Most research has focused on English, but we include results of  recent models which were evaluated on at least 3 languages.
When compared to the previous models, both our models are very competitive, with the exception of German. On the German dataset, \newcite{mulcaire-etal-2018-polyglot} also report  a relatively weak result, when compared to  \newcite{roth-lapata-2016-neural}. The German dataset is the smallest one in terms of the number of predicates.  
 Syntactic information used by
\newcite{roth-lapata-2016-neural} may be very beneficial in this setting and may be the reason for this discrepancy.
Our structured refinement approach improves over the best previous results on 5 out of 7 languages. 
Note that  hyper-parameters of the refinement network are not tuned for individual languages, suggesting that the proposed method is robust
and may be easy to apply to new languages and/or new base models.
The only case where the refinement network was not effective is Chinese, where it achieved only a negligible improvement. 

\begin{table}[ht!]
\centering
 \begin{tabular}{ c c c} 
 \hline
 English &  Test & Ood \\ [0.5ex] 
 \hline
\newcite{Li2018DependencyOS}& 90.4 & 81.5 \\ 
 \hline
 Baseline   &  90.65  & 81.98  \\
Structured Refinement   &    \bf90.99  & \bf82.18  \\
 \hline
 German  & Test & Ood \\ [0.5ex] 
 \hline
\newcite{Zhao2009MultilingualDL} & \bf 76.19 & \bf67.78 \\ 
 \hline
 Baseline  & 75.06  & 65.25 \\
Structured Refinement   & 75.87  & 65.69  \\
 \hline
 Czech & Test & Ood \\ [0.5ex] 
 \hline
 \newcite{Marcheggiani2017ASA} & 86.0 &  \bf 87.2 \\ 
 \hline
 Baseline   &  87.30  &   85.80  \\
Structured Refinement  &  \bf  87.62   & 86.04 \\
 \hline
 \end{tabular}
\caption {Labeled F1 scores (including senses) on English, German, Czech in-domain and out-of-domain test sets; we chose the previous models achieving the best scores on the out-of-domain test sets.}
 \label{tb:out}
\end{table}

\begin{table*}[h]
\centering
 \begin{tabular}{c| c c c c c c c | c} 
 \hline
 Model &Ca &Cz& De& En& Ja & Es & Zh & Avg.\\ [0.5ex] 
 \hline
 Baseline Model &  80.69   & 87.30  &  75.06   &      90.65  &    81.97 & 79.87 & 83.26  & 82.69\\
 Self Refinement  &80.65    & 87.32 &    74.83&    90.71&    82.27&    80.08&    \bf 83.32 & 82.74 \\
 Structured Refinement  & \bf 80.91    & \bf 87.47&    \bf 75.83& \bf    90.83&    \bf 82.54&\bf     80.53&    83.31 & \bf 83.06\\
 \hline
 \end{tabular}
\caption {Labeled F1 score (including senses) for all languages on the CoNLL-2009 in-domain test set. }
 \label{tb:ab_struc}
\end{table*}

\noindent{\bf Out-of-Domain} Results on the out-of-domain testing sets are presented in Table~\ref{tb:out}.\footnote{\newcite{roth-lapata-2016-neural} has better in-domain testing score, but did not report the out-of-domain score.} We observe improvements from using refinement in all the cases. This shows that our refinement approach is robust against domain shift. 

\noindent{\bf Ablations} We report development set results in different settings in Table~\ref{tb:ab1}.
Our full model performs 2 refinement iterations, uses weight tying, and the Gumbel noise.\footnote{We set  $\lambda^\alpha_g = 5$ for role and $\lambda^\pi_g = 50$ for sense, so that initial predictions contain around 20\% errors.} We select the best configuration for each language to report the test set performance in Table~\ref{tb:mul} and Table~\ref{tb:out}. 

As expected, weight tying is beneficial for lower-resource languages such as Catalan, Japanese and Spanish (see Table~\ref{tb:stats} for dataset characteristics). 
The  Gumbel noise helps for all languages except for Czech and English, the two largest datasets. In particular, we observe almost no improvement on the Spanish dataset without using the Gumbel noise. We note relatively consistent but small gains from using 2 refinement iterations.
The magnitude of the gains may be an artifact of us having the loss terms $\mathcal{L}(\mathrm{R}^*,\mathrm{R}^t)$ and $\mathcal{L}(\mathrm{P}^*,\mathrm{P}^t)$, encouraging not only the final (second), but also the first, iteration to produce accurate predictions.
A potential  alternative explanation is that our  refinement network is restricted to simple interactions, resulting in the fixed point reachable in one step.

\begin{table}[ht]
\centering
\begin{tabular}{c |  c c c}
\toprule
Model &  U & C & R\\
 \hline
Gold   & 55 & 0  & 88 \\
Baseline   & 301 &2 &114 \\
Structured Refinement   &  142 & 2  & 111 \\
\bottomrule
\end{tabular}
\caption{Unique core roles violations (U), continuation roles violations (C) and reference roles violations (R) on English in-domain test set.}
\label{tb:vio}
\end{table}
\noindent{\bf Constraints Violation } We consider violation of unique core roles (U), continuation roles (C) and reference roles (R) constraints from~\newcite{Punyakanok:2008:ISP:1403157.1403162,FitzGerald2015SemanticRL}  in Table~\ref{tb:vio}. U is violated if a core role (A0 - A5, AA) appears more than once; 
C is violated when the C-X role is not preceded by the X role (for some X); R is violated if R-X role does not appear. 
Our approach results in a large reduction in the uniqueness constraint violations. Our model slightly reduces the number of R violations, while \newcite{P17-1044} reported that deterministically enforcing constraints is not helpful (albeit in span-based SRL). However learning those constraints in a soft way might be beneficial. 

\noindent{\bf Argument Interaction vs. No Argument Interaction}  We compare the structured refinement network and the self-refinement network in Table~\ref{tb:ab_struc}. Both networks share the same hyper-parameters.
The structured refinement network consistently outperforms the self-refinement counterpart. 
This suggests that the refinement model benefits from accessing information about other arguments when doing refinement. In other words, modeling argument interaction appears genuinely useful. 
\begin{table}[ht]
\begin{tabular}{c | c  c c}
 \hline
Model & RP & RR  & Sense\\
 \hline
Baseline   & 88.1 & 88.3     & 96.2\\
Structured Refinement  &  88.7 & 88.5   & 96.3\\
 \hline
\end{tabular}
\caption{Labeled roles precision (RP), recall (RR) and sense disambiguation accuracy (Sense) on English in-domain test set.}
\label{tb:decompose}
\end{table}

\noindent{\bf Improvement Decomposition} We report labeled role precision, recall and sense disambiguation accuracy in Table~\ref{tb:decompose}. 
Our structured refinement approach consistently improves over the baseline model in all metrics. While we cannot assert the improvements on all metrics are significant, this suggests that it learns some non-trivial interactions instead of merely learning to balance precision and recall.

\begin{figure}[ht]
\centering
\includegraphics[width=1\columnwidth]{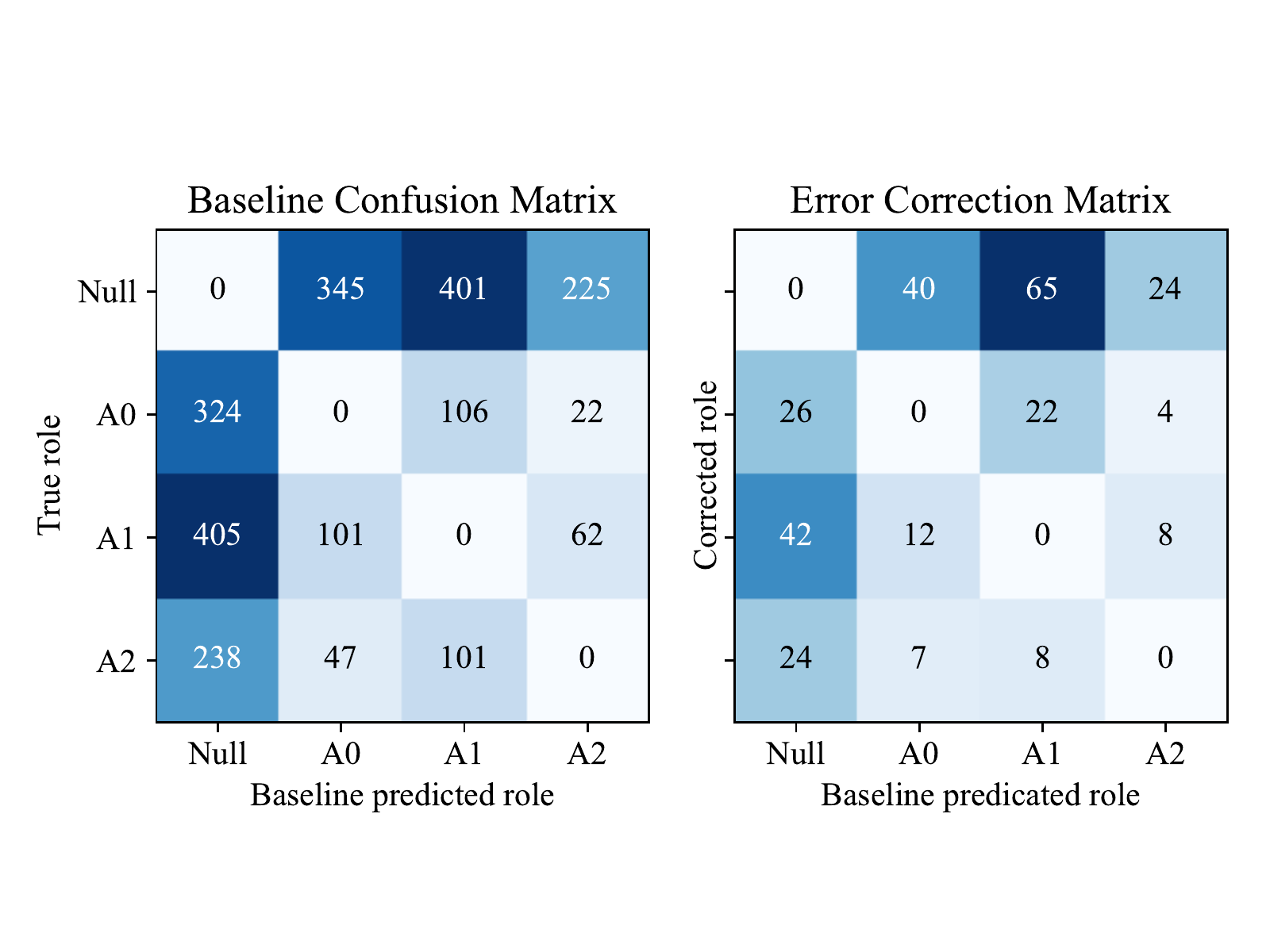}
\caption{Confusion matrix for the baseline model, and a correction matrix where the errors were corrected by the refinement network. Only Null, A0, A1, A2 are presented here.}
\label{fig:correct}
\end{figure}

\noindent{\bf Error Correction Analysis } We show the errors that the structured refinement network corrects in Figure~\ref{fig:correct}. In the baseline confusion matrix, we see the errors are fairly balanced for all the roles we consider here. In the error correction matrix, the corrections are also fairly evenly distributed. Yet, this is not completely uniform. There is a tendency towards filtering out arguments rather than generating new ones. 


\section{Conclusions and Future Work}
We propose the structured refinement network for dependency semantic role labeling. The structured refinement network corrects predictions made by a strong factorized baseline model while modeling interactions in the predicated structure. 
The resulting model achieves state-of-the-art results on 5 out of 7 languages in the CoNLL-2009 data set, and
substantially outperforms the factorized model on all of these languages.

For the future work, the structured refinement network can be further improved. For example, we can take an inspiration from either declarative constraints used in the previous work~\cite{Punyakanok:2008:ISP:1403157.1403162} or from literature on lexical semantics of verbs, studying patterns of event and argument realization~(e.g., \citealt{levin1993english}). Indeed, the unique role constraint as a declarative constraint is one of the motivation for the concurrent work on modeling argument interaction in SRL~\cite{CapsuleRefine}. That work
relies on capsule networks~\cite{sabour2017dynamic} and
 focuses primarily on enforcing the role uniqueness constraint.

The framework can be extended to other tasks. For example, in syntactic dependency parsing: the refinement network can rely on representations of grandparent nodes, siblings and children to propose a correction. In general, structure refinement networks should allow domain experts to incorporate prior knowledge about output dependencies and improve model performance.


\section*{Acknowledgments}
We thank the anonymous reviewers for their suggestions. The project was supported by the European Research Council (ERC StG BroadSem 678254), the Dutch National Science Foundation (NWO VIDI 639.022.518) and Bloomberg L.P.
\nocite{Srivastava2015TrainingVD,Clevert2015FastAA,Gal2016ATG}
\bibliography{emnlp-ijcnlp-2019}
\bibliographystyle{acl_natbib}

\onecolumn 
\section*{\LARGE\centering{Supplementary Material }}
\vspace{4cm}

 \subsection*{Variables Summary}
\begin{table}[H]
\centering
\begin{tabular}{c  | c | c}
 \hline
Variable &  Domain & Description\\
 \hline
$n$  & $\mathbb{N}_+$& sentence length \\
$r$  & $\mathbb{N}_+$ & number of semantic roles \\
$m$  & $\mathbb{N}_+$ & number of senses  \\
$T$  & $\mathbb{N}_+$ & number of refinement iterations \\
$j$  & $\mathbb{N}_+$ & position of the given predicate \\
$\mathrm{R}$  & $\Delta^{n}_{ r-1}$ & semantic roles probability\\
$\mathrm{P}$  & $\Delta_{m-1}$ & senses probability\\
$\Pi$  & $\mathbb{R}^{m\times d_\pi}$ & predicate senses embedding  \\
$x^\mathrm{w}$  & $\mathbb{R}^{n\times d_w}$ & sentence tokens embeddings\\
$x^\mathrm{dep}$  & $\mathbb{R}^{n\times d_\delta}$ &   dependency labels embeddings\\
$x^\mathrm{pos}$  & $\mathbb{R}^{n\times d_{p}}$ &  part-of-speech tags embeddings\\
$x$  & $\mathbb{R}^{n\times (d_{w}+d_\pi + d_p )}$ & concatenated representation  \\
\hline
Baseline Model & &\\
$\mathrm{h}$  & $\mathbb{R}^{n\times d_h}$ & encoded sentence representation  \\
$\mathrm{h}^{\rho_0}$  & $\mathbb{R}^{n\times d_{\rho_0}}$ & argument feature for null role logits  \\
$\mathrm{h}^{\rho_1}$  & $\mathbb{R}^{n\times d_{\rho_1}}$ & argument feature for other roles logits  \\
$\mathrm{h}^{\varrho_0}$  & $\mathbb{R}^{ d_{\rho_0}}$ & predicate feature for null role logits  \\
$\mathrm{h}^{\varrho_1}$  & $\mathbb{R}^{d_{\rho_1}}$ & predicate feature for other roles logits  \\
$\mathrm{I}^{\rho_0}$  & $\mathbb{R}^{n}$ &  null role logits  \\
$\mathrm{I}^{\rho_1}$  & $\mathbb{R}^{n\times (r-1)}$ &other roles logits  \\
$\mathrm{I}^{\alpha}$  & $\mathbb{R}^{n\times r}$ & roles logits  \\
$\mathrm{h}^{\pi}$  & $\mathbb{R}^{d_{\pi}}$ & predicate feature for sense disambiguation  \\
$\mathrm{I}^{\pi}$  & $\mathbb{R}^{m}$  & sense logits  \\
\hline
Refinement Network & &\\
$\mathrm{g}$  & $\mathbb{R}^{n\times d_g}$ & encoded sentence representation  \\
$\mathrm{g}^{\alpha}$  & $\mathbb{R}^{n\times d_{\alpha}}$ & argument feature   \\
$\mathrm{g}^{\pi}$  & $\mathbb{R}^{d_{\pi}}$ & predicate feature   \\
$\mathrm{o}_i$  & $\mathbb{R}^{r-1}$ & sum of other roles   \\
$\mathrm{z}^\alpha_i$  & $\mathbb{R}^{2r-1+2d_g+d_\pi}$ & input to role refinement network  \\
$\mathrm{M}^\alpha$  & $\mathbb{R}^{n \times r}$ & refinement role logits to be added with $\mathrm{I}^{\alpha}$\\
$\mathrm{r}_\pi$  & $\mathbb{R}^{r-1}$ & sum of all roles   \\
$\mathrm{z}^\pi$  & $\mathbb{R}^{2r-1+2d_g+d_\pi}$ & input to sense refinement network  \\
$\mathrm{M}^\pi$  & $\mathbb{R}^{m}$ & refinement sense logits to be added with $\mathrm{I}^{\pi}$\\
 \hline
\end{tabular}
\caption{Variables domain and description}
\label{tb:var}
\end{table}

 \subsection*{Networks}
\begin{table}[H]
\begin{tabular}{c | c }
\toprule
Network  & {Description}\\
 \hline
 $ \text{BiLSTM} $& 3 layers stacked highway BiLSTM\\
 $\text{MLP}  $&  1 layer MLP with exponential linear units\\
\bottomrule
\end{tabular}
\caption{Networks for all languages at all occurence in the main text. Note that the input and output dimensions of $\text{MLP}$ and $\text{BiLSTM}$ can be decided by the other hyper-parameters at each occurrence.}
\label{tb:net}
\end{table}

 \subsection*{Hyper-Parameters}
\begin{table}[H]
\begin{tabular}{c | c c c | c}
\toprule
Hyper-parameter  &  \multicolumn{3}{c}{Value} & \multicolumn{1}{c}{Description}\\
\midrule 
{}   & English  & & Others      & \\
 \hline
 $p$ & & 0.3 & & dropout rate for all neural modules \\
 $p_r$ & & 0.3 & & recurrent dropout rate for BiLSTMs  \\
$d_w$  & 1024 &  & 300 &  tokens embedding dimension\\
$d_\delta$  &  & 64 &  &  dependency label embedding dimension\\
$d_p$  &  & 64 &  &  part-of-speech tags embedding dimension\\
$d_{h}$  & 500 &  & 428 &  BiLSTM hidden state dimension in one direction\\
$d_{\rho_0}$  &  & 300 &  &  dimension for feature for null role logits \\
$d_{\rho_1}$  &  & 128 &  &  dimension for feature for other role logits \\
$d_{g}$  &  & 200 &  &  dimension for feature for refinement networks \\
$d_{r}$  &  & 200 &  &  hidden dimension of refinement networks \\
$\lambda_g^\pi$  &  & 50 &  &  multiplier of Gumbel noise for sense logits  \\
$\lambda_g^\alpha $ &  & 5 &  &  multiplier of Gumbel noise for role logits  \\
\bottomrule
\end{tabular}
\caption{Hyper-parameters value and description. Note that the input and output dimensions of $\text{MLP}$ and $\text{BiLSTM}$ can be decided by the other hyper-parameters at each occurrence.}
\label{tb:hyper}
\end{table}

\end{document}